\documentclass[sigconf]{acmart}

\usepackage{booktabs} 
\usepackage[ruled, boxed, linesnumbered, commentsnumbered, vlined]{algorithm2e}
\usepackage{flushend}
\usepackage{float}

\copyrightyear{2018}
\acmYear{2018}
\setcopyright{acmcopyright}
\acmConference[ICVGIP 2018]{11th Indian Conference on Computer Vision, Graphics and Image Processing}{December 18--22, 2018}{Hyderabad, India}
\acmBooktitle{11th Indian Conference on Computer Vision, Graphics and Image Processing (ICVGIP 2018), December 18--22, 2018, Hyderabad, India}
\acmPrice{15.00}
\acmDOI{10.1145/3293353.3293383}
\acmISBN{978-1-4503-6615-1/18/12}

\begin{document}
\title[HSD-CNN using class specific filter sensitivity analysis.]{HSD-CNN: Hierarchically self decomposing CNN architecture using class specific filter sensitivity analysis}
\titlenote{We are extremely thankful to Tom Michiels and Bert Moons from Synopsys, Leuven, Belgium who provided technical expertise that greatly assisted in the proposed, and also for the comments that greatly improved the manuscript.}

\author{K. SaiRam}
\affiliation{%
  \institution{Indian Institute of Technology Kharagpur}
}
\email{sairam.kasanagottu@gmail.com}

\author{Jayanta Mukherjee}
\affiliation{%
  \institution{Indian Institute of Technology Kharagpur}
}
\email{jay@cse.iitkgp.ac.in}

\author{Amit Patra}
\affiliation{
  \institution{Indian Institute of Technology Kharagpur}
  }
\email{amit@ee.iitkgp.ernet.in}

\author{Partha Pratim Das}
\affiliation{%
  \institution{Indian Institute of Technology Kharagpur}
  }
\email{ppd@cse.iitkgp.ernet.in}



\renewcommand{\shortauthors}{K. SaiRam et al.}

\begin{abstract}
Conventional Convolutional neural networks (CNN) are trained on large domain datasets and are hence typically over-represented and inefficient in limited class applications. An efficient way to convert such large many-class pre-trained networks into small few-class networks is through a hierarchical decomposition of its feature maps. To alleviate this issue, we propose an automated framework for such decomposition in Hierarchically Self Decomposing CNN (HSD-CNN), in four steps. HSD-CNN is derived automatically using a class-specific filter sensitivity analysis that quantifies the impact of specific features on a class prediction. The decomposed hierarchical network can be utilized and deployed directly to obtain sub-networks for a subset of classes, and it is shown to perform better without the requirement of retraining these sub-networks. Experimental results show that HSD-CNN generally does not degrade accuracy if the full set of classes are used. Interestingly, when operating on known subsets of classes, HSD-CNN has an improvement in accuracy with a much smaller model size, requiring much fewer operations. HSD-CNN flow is verified on the CIFAR10, CIFAR100 and CALTECH101 data sets. We report accuracies up to $85.6\%$ ( $94.75\%$ ) on scenarios with 13 ( 4 ) classes of CIFAR100, using a pre-trained VGG-16 network on the full data set. In this case, the proposed HSD-CNN requires $3.97 \times$ fewer parameters and has $71.22\%$ savings in operations, in comparison to baseline VGG-16 containing features for all 100 classes.

\end{abstract}



\maketitle

\section{Introduction} \label{s:intro}

Recently, Convolutional Neural Networks ( CNNs ) have outperformed traditional machine learning models in many computer vision tasks. However, it required extensive research in discovering high-performance CNN architectures. As far as the large-scale image classification task is concerned, state-of-art CNN's are going beyond deep, and single chain structured layouts \cite{DBLP:journals/corr/SimonyanZ14a, 7780459}.

All these networks are trained on datasets with many classes and are over-represented when they are used on smaller tasks with fewer classes \cite{DBLP:journals/corr/SimonyanZ14a}. This over-representation translates into large and inefficient models, that require too many weights to represent redundant features and too many computations to compute. Instead of a single chain structured CNN, a hierarchically structured CNN would be more efficient, as it allows using only the necessary features to represent a specific subset of classes, rather than the full set of features used to represent the full class domain set.

So, our objective is to design a network that handles large classes and simultaneously inhibits the over-representation between the classes with minimal manual interference and design time. Though there are 2-level hierarchical strategies exploited in \cite{yanhd,  DBLP:journals/corr/abs-1802-05800, Zhu2017BCNNBC}, methods to deploy model for classifying specific classes without retraining the network are not found. So, we adopt filter sensitivity analysis in \cite{DBLP:conf/cvpr/GuoP17}, and form Impact score class vectors (\textit{Iscv}). \textit{Iscv}s also helps in automated designing of network architecture.

In this paper, we propose an automated way to create computationally efficient and compressed Hierarchically Self Decomposed CNN's (HSD-CNN), based on existing pre-trained models. In the proposed algorithm, classes are organized hierarchically without manual intervention. The automated design flow of an HSD-CNN is a four-step process, detailed as in Section \ref{s:HSD}. Part of HSD-CNN that discriminates specific set of classes, named as Subnetwork, is selected in achieving our objective, and the corresponding sub-model can be deployed without retraining, detailed in Section \ref{sss:subnet}.

This paper has two main contributions:

\begin{itemize}
	\item Any state of the art CNN's can be automatically decomposed hierarchically into multiple levels using class specific filter analysis \cite{DBLP:conf/cvpr/GuoP17}. This is the first time a CNN is decomposed and pruned simultaneously based on class-specific filters. HSD-CNN algorithm allows the discriminating learned features to be organized hierarchically. These features are not limited to just two levels - coarse and fine as in \cite{yanhd, DBLP:journals/corr/abs-1802-05800, Zhu2017BCNNBC}. HSD-CNN has more than two layers, where classes can be grouped hierarchically.
	\item Sub-network of HSD-CNN can be used to overcome over-represented CNN models. Part of HSD-CNN corresponding to a specific class domain are adapted as efficient subnetworks. Subnetworks are deployed without retraining to any application scenarios where only a subset of classes is used. And, results show that sub-networks perform better for most cases by a good margin.
\end{itemize}


\section{Recent Literature} \label{s:lit}
With an increase in the complexity of architecture, high computation and memory requirements of these models hinder their deployment on low power embedded devices. So far, many researchers have focused on pruning individual parameters on powerful models, at the cost of performance degradation. Le Cun et al. \cite{Cun:1990:OBD:109230.109298} analytically prunes those parameters that have fewer effects when these parameters are perturbed. In \cite{Hassibi:1992:SOD:645753.668069}, second order derivatives on loss function are used to determine the parameters which need to be pruned. Han et al. \cite{Han:2015:LBW:2969239.2969366} achieves impressive memory savings by removing weights with magnitudes smaller than a threshold. In \cite{Anwar:2017:SPD:3051701.3005348, 7303876}, filters and parameters are pruned at different levels of the model using statistical analysis of filters and their feature maps. Method of pruning individual parameters and filters have resulted mostly in compressing the model, but not speeding up the inference time.

As filters at each layer in CNNs are tensors or matrices ( tensor slices ), applying low rank approximation methods \cite{Denton:2014:ELS:2968826.2968968,DBLP:journals/corr/JaderbergVZ14,8099688,Zhang:2016:AVD:3026801.3026837,DBLP:journals/corr/HeZS17,DBLP:conf/iccv/WenXWWCL17,8354187} to decompose these filters into lightweight layers have been useful in increasing the efficiency during inference, and simultaneously reducing the number of parameters. Design methodolgy found in \cite{Denton:2014:ELS:2968826.2968968,DBLP:journals/corr/JaderbergVZ14,8099688,Zhang:2016:AVD:3026801.3026837,DBLP:journals/corr/HeZS17,DBLP:conf/iccv/WenXWWCL17,8354187} lead to discovering compact and efficient networks like MobileNet \cite{DBLP:journals/corr/HowardZCKWWAA17}, SqueezeNet \cite{DBLP:journals/corr/IandolaMAHDK16}, Grouped Convolutions \cite{DBLP:journals/corr/ZhangZLS17, Huang_Shichen_VanderMaaten_Weinberger_2018}. 

Very few attempts have been made to exploit category hierarchies \cite{yanhd,  DBLP:journals/corr/abs-1802-05800, Zhu2017BCNNBC} in deep CNN models, and they are restricted to two levels. Hierarchical deep CNN's (HD-CNN) \cite{yanhd} embed deep CNN's into a two-level category hierarchy. They separate easy classes using a coarse category classifier while distinguishing difficult classes using fine category classifiers. Tree-CNN \cite{DBLP:journals/corr/abs-1802-05800} proposes a training method for incremental learning, albeit the network is limited to similar 2-level hierarchy as in \cite{yanhd}. A similar 2-level hierarchy along with Branch Training strategy is introduced in Branch-CNN \cite{Zhu2017BCNNBC}. The branch training strategy balances the strictness of the prior with the freedom to adjust parameters on the output layers to minimize the loss.

Distributed representations in the hidden layers of deep feed-forward neural networks have excellent generalization abilities \cite{DBLP:journals/nature/LeCunBH15}, though these representations are difficult to analyze. Because any particular feature activation depends on the effects of all other units in the same layer in its distributed representation.

Despite the limited understanding of the hidden representation that discriminates the class, in \cite{DBLP:journals/nature/LeCunBH15}, the authors of \cite{DBLP:journals/corr/abs-1711-09784} proposes mimicking a neural net as a decision tree that makes soft decisions. This method allows to form a short representation as a decision tree and with faster execution. However, they do not explore distilling the CNNs as a decision tree. Because CNNs have a lot of information which cannot be represented in the form of a decision tree. Information will be lost if CNNs are represented as a normal decision tree. So, we require a method to represent CNNs as a decision tree, without losing information.

Authors of \cite{DBLP:journals/corr/abs-1802-00121} explain that logic behind each prediction made by a pre-trained CNN can be quantitatively represented by a decision tree. It also explains that each filter of a layer might represent a specific or group of object parts. But, authors present no strategy to influence inference computation and model compression. So, pruning based on model interpretability is still a significant challenge in neural networks ( NNs ).

Authors of \cite{DBLP:conf/cvpr/GuoP17} proposes a filter sensitivity analysis method to decide the filter importance specific to a class. We adopt this method in interpreting the trained model of a network that can handle a large number of classes. So, the proposed network decomposes itself based on the interpretation computed earlier using class specific filter sensitivity analysis.  And, the decomposed network structure depicts a decision tree when observed in its computational graph analogy form.

\section{Hierarchically Self Decomposing CNN}  \label{s:HSD}

\subsection{Preliminaries} \label{ss:prelim}
Consider a classical image classification setting on a conventional CNN learned through a training dataset $\mathbb{D}$ over a set of classes $\mathbb{C}$. The dataset $\mathbb{D}$ is composed of samples of images and corresponding labels $ (X_i, y_i), i = 1,2,3, \cdots, \lvert\mathbb{D}\rvert$, where $X$ is a 3-dimensional input image and $y$ is associated with one of the class labels in $\mathbb{C}$.


The goal of the CNN in image classification is to learn a mapping function $y = f(X)$. Softmax function at the final layer of the network produces a posterior distribution over classes $\mathbb{C}$. Then the network minimizes a classification loss function over all samples of dataset $\mathbb{D}$ to search optimal parameters required for the mapping function.

\begin{equation} \label{eq:minloss}
    \min \sum_{i=1}^{\lvert\mathbb{D}\rvert} \mathcal{L}(f(X_i),y_i)
\end{equation}
Surprisingly, any modification in Equation [\ref{eq:minloss}] formation affects the performance of the assumed image classification setting. For example, deep CNNs \cite{DBLP:journals/corr/SimonyanZ14a}, with many layers, performs better for a large scale image scale classification.

Yet, optimizing the loss function in Eq. [\ref{eq:minloss}] is troublesome due to the curse of dimensionality, i.e, increase in complexity of the network and its parameters. Designing new networks catering to the scenario of a large number of classes is a time-consuming and tedious task. Using large networks for application domains that are restricted to a subset of its classes $\mathbb{C}$ is an overkill, as the network is over-represented in chain structured conventional CNN. Because, the learned discriminative features may be more biased towards a specific set of classes, and may worsen the performance (like accuracy, inference, latency speed) on the remaining class set.

\begin{figure}[t]
\includegraphics[width=0.4\textwidth,height=10cm,keepaspectratio]{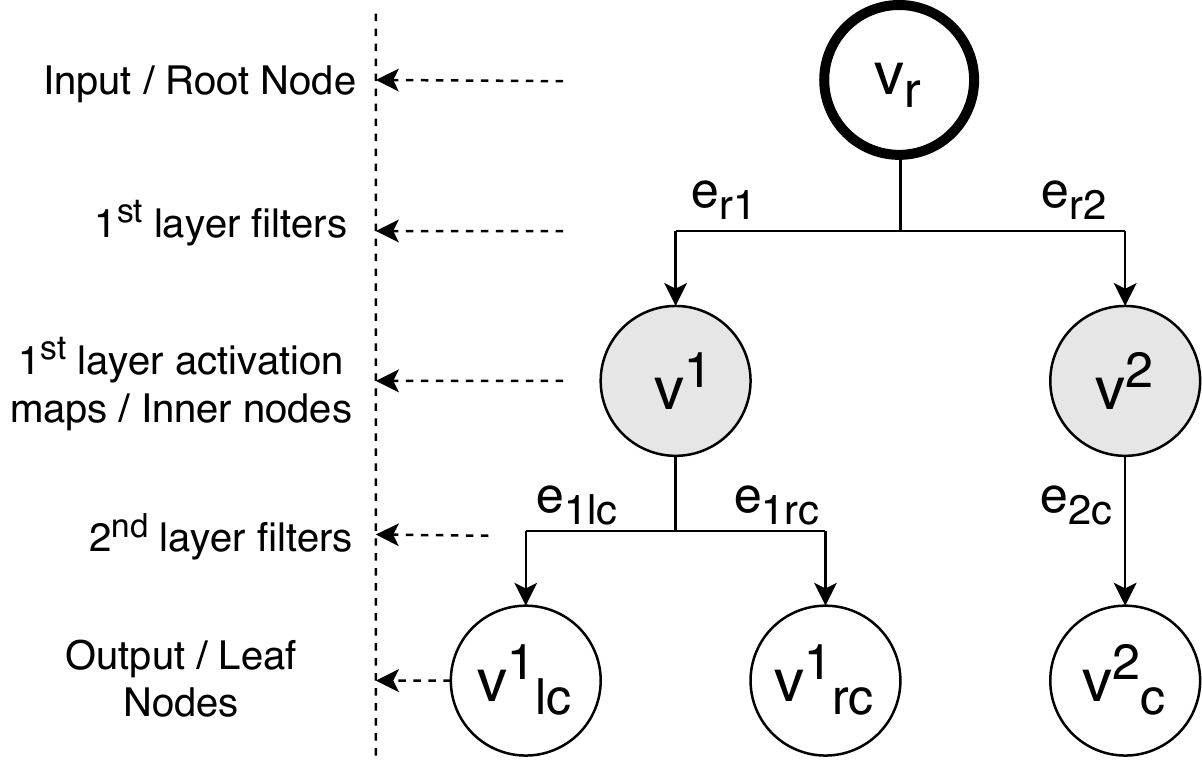}
\caption{CNN as DAG, $G=(V,E)$, where $V=\{v_r, v^1, v^2, v^1_{lc}, v^1_{rc}, v^2_c\}$ are activation map nodes, and $E=\{e_{r1}, e_{r2}, e_{1lc}, e_{1rc}, e_{2c}\}$ are the filters associated in their layers. Here, inner nodes $v^1,v^2$ are marked in \textit{gray}, root node $v_r$ as \textit{thick black border}, and leaf nodes $V_O=\{v^1_{lc}, v^1_{rc}, v^2_c\}$ are in \textit{plain} color. By parent node representation, $P(v^1) = P(v^2) = v_r$, $P(v^1_{lc}) = P(v^1_{rc}) = v^1$, and $P(v^2_c) = v^2$. $v^1_{lc}, v^1_{rc}$ are left and right child nodes of $v^1$, and $v^2_c$ is single child node of $v^2$  }
\label{fig:dag}
\end{figure}

So, our objective is to design a network that simultaneously handles large classes and inhibits the over-representation between the classes with minimal manual interference and design time. For easier understanding of our objective, it is useful to represent CNN as a directed acyclic graph (DAG), $G=(V, E)$, where $V,~E$ is the set of nodes and edges in network graph $G$, respectively. Visual representation of network graph G is in Figure \ref{fig:dag}.

Let us also represent $P(v)$ as the parent node of node $v$, for all nodes except root node in $V$. Root node $v_r$ is the input image and leaf node $v_o$ is the probability score over classes $\mathbb{C}$. Also, let us refer all nodes except root node and leaf nodes as inner nodes, which represent activation maps.

We also represent the number of edges from the root node $v_r$ to any node $v$ as $l$. It signifies that the node $v$ is present at layer $l$ in network $G$. Also, network $G$ at layer $l$ may have more than one nodes with same or different parent nodes. Layer definition is similar to the definition of depth in graphs. 


Node $v$ is a 3-dimensional activation map. The output node $v$ is the convolution output formed between input activation map $P(v)$ and  $l^{th}$ layer filters in edge $e(P(v), v)$. Generally, each edge e has $K$ 3-dimensional filters which may vary with edges in $E$.

Node $v$ is useful in discriminating classes  $\mathcal{C}(v)$ - a subset of classes $\mathbb{C}$ in dataset $\mathbb{D}$. For any inner node node $v$, $\mathcal{C}(v) = \mathcal{C}(v_{lc}) \cup \mathcal{C}(v_{rc})$. We also restrict overlapping class subsets, $\mathcal{C}(v_{lc}) \cap \mathcal{C}(v_{rc}) = \phi$. So, $\mathcal{C}(v_r) = \cup_{V_O} \mathcal{C}(v_o) = \mathbb{C}$, and $\cap_{V_O} \mathcal{C}(v_{o}) = \phi$. 

In a conventional CNN $G^C = (V^C, E^C)$, each node $v \in V^C$ has one child $v_c$, except for leaf nodes $v_o$. Number of CONV layers in conventional network from input image to present layer activation map is counted as $l$. 

Similarly, in HSD-CNN $G^H = (V^H,E^H)$, each inner node $v \in V^H - \{v_r^H, V_O^H\}$ is restricted to at least one child node ( $v_{c}$ ) and at most two children nodes ( $v_{lc}, v_{rc}$ ).

\subsection{Algorithm} \label{ss:algo}
Our algorithm proposes self-decomposition of conventional chain-structured CNN models into a tree-structured CNN layout ( HSD-CNN ). The proposed inherently categorizes whole classes of $\mathbb{C}$ into a hierarchical group of subsets in the following four steps:

\subsubsection{\textbf{Impact score class vector}}\label{sss:icsv}

Let us analyze the impact of a channel on a certain class at layer $l$ in the network. It requires a large number of variables to examine the channels are inter-dependent on its predecessor and successive layer features. Our proposed method adapts filter sensitivity analysis \cite{DBLP:conf/cvpr/GuoP17} to calculate the impact of each channel in layers on each class present in the dataset $\mathbb{D}$ as a score.

Let the trained conventional CNN be $G^C$. Assign a weight $w_k^l = 1$ for each channel $k$ at  layer $l$ in the network $G^C$, with no changes to other layers. Let the modified network be $G^{M}$. Suppose an $i^{th}$ sample $X_i$, with corresponding ground truth class $c$, in the training data produces a feature map $v^C = \{x_{ik}^l\}, k = 1,2,3, \cdots, K$ at layer $l$ in network $G^C$. Its response at corresponding $l^{th}$ layer of $G^{M}$ results in $v^{M} = \{w_{k}^l~.~x_{ik}^l\}, k = 1,2,3, \cdots, K$ as the weight variable $w_k^l$ is included in $G^{M}$. 

Let $p_{ci}$, an element of leaf node vector $v_o$ in $G^{M}$ be the probability score corresponding to the class $c$ that the sample $X_i$ belongs to. It is calculated from the softmax layer output.

Then the impact score $I_{ikc}^l$ that the channel $k$ at $l^{th}$ layer node $v$ in $G^{M}$ has on class $c$ is defined as ratio of the change in the probability score with effective change in weight variable $w_k^l$ at corresponding $k^{th}$ channel.

\begin{equation}
    I_{ikc}^l = \frac{\delta p_{ci}}{\delta w_{k}^l}
\end{equation}

For inclusion of robustness in the impact score $I_{ikc}^l$, let us calculate the sum of absolute values of the impact scores produced for each sample $X_i$ whose class label is $c$.

\begin{equation} \label{e:imps}
    I_{kc}^l = \sum_{X_i|y_i = c}{\left\lvert\frac{\delta p_{ci}}{\delta w_k^l}\right\rvert}
\end{equation}

As there are $K$ channels to $l^{th}$ layer node $v$, there are $K$ impact scores for each class $c$. Let us represent all the scores at particular layer $l$ for a class as a feature vector, namely Impact score class vector, $\widehat{Iscv}_c^l = \{I_{kc}^l\}_{1XK}, k = 1,2,3,\cdots, K$. Lastly, we normalize the vector by dividing with its maximum element $I_{k_{max}c}$, i.e, $\textit{Iscv}_c^l = \{I_{kc}^l/I^l_{k_{max}c}\}_{1XK}, k = 1,2,3,\cdots, K$ and $k_{max}$ is the index of maximum element. It helps in bringing vectors $\textit{Iscv}_c^l$ in a convenient range for comparison with other class vectors. Normalization of \textit{Iscv} mitigates minimum-maximum variation values for different classes. Calculation of \textit{Iscv} features is repeated for all classes at each layer in $G^C$. These \textit{Iscv}s help in self formation of our tree-structured CNN, $G^H$.

\subsubsection{\textbf{Formation of HSD-CNN}} \label{sss:form}

Generally, a decision tree forms from a supervised algorithm that classifies data based on a hierarchy of rules learned over the training samples. Each internal node in the tree represents an attribute, while each leaf node represents a decision on the input sample. To build a tree, we start with a root node. Similarly, HSD-CNN $G^H$ is formed by self-decomposition of nodes from conventional CNN $G^C$. $G^H$ starts with a root node, which is an input sample.

\begin{algorithm}[h]
\SetAlgoLined
\KwResult{Nodes $V^H$ and its empty edges $E^H$ in $G^H$.}
 initialization: Empty $G^H$\;
 $v_r^H \gets v_r^C$\;
 $insert ~ node~ v_r^H \in G^H$\;
 $queue = [v_r^H]$\;
 \While{queue}{
  $v^H \gets$ queue.pop()\;
  $l \gets L(v^H)$\;
  $v_c^C  \gets $ child node of $v^C$ at $l^{th}$ layer of $G^C$\;
  \If{$no~ child~ v_c^C$}{
   \textit{continue}\;
   }
  $v_{lc}^H, ~ v_{rc}^H ~= \textit{DecomposeNode(Iscv}^l(v_c^C), \mathcal{C}(v^H))$\;
  $\textit{assert} ~~ \mathcal{C}(v_{lc}^H) \cup \mathcal{C}(v_{rc}^H) == \mathcal{C}(v^H)$\;
  $assert ~~ \mathcal{C}(v_{lc}^H) \cap \mathcal{C}(v_{rc}^H) == \phi$\;
  \If{$v_{lc}^H$}{
   queue.insert($v_{lc}^H$)\;
   $insert~ node~~ v_{lc}^H \in G^H$\;
   $insert~ empty~filter~e(v^H,v_{lc}^H) \in G^H$\;
   $assert ~~ P(v_{lc}^H) == v^H $\;
   }
   \If{$v_{rc}^H$}{
   queue.insert($v_{rc}^H$)\;
   $insert~ node~ v_{rc}^H \in G^H$\;
   $insert~ empty~filter~e(v^H,v_{rc}^H) \in G^H$\;
   $assert ~~ P(v_{rc}^H) == v^H $\;
   }
 }
 \caption{Formation of HSD-CNN layout $G^H$, given $G^C=\{V^C,E^C\}$ and \textit{Iscv}}
 \label{al:tree}
 
\end{algorithm}

Let the layer for a node $v$ in $G$ given by $L(v)$.

As in Algorithm \ref{al:tree}, HSD-CNN tree graph $G^H$ is initialized with a root node, along with queue $Q$. For each node $v^H$ out of queue $Q$, select node $v^C$ from $G^C$ at $L(v^H)$ layer along with its child node $v_c^C$. As the $G^C$ is a single chain structured CNN, only one $v^C$ and one $v_c^C$ is available.

\textit{Decompose Node} module of the algorithm either clusters the classes $\mathcal{C}(v^H)$ into two subsets or one set, and discussed in details further. For a given node $v_c^C$ and its \textit{Iscv}s, \textit{Decompose Node} results in either one or two nodes $v_{lc}^H, ~v_{rc}^H$, which are attached as child nodes to $v_H$ with an empty edge $e(v^H, v_{l/rc}^H)$ in $G^H$. Correspondingly, these nodes are inserted in a queue to repeat the steps till the queue is empty. If it results in only one child, then $\mathcal{C}(v_{rc}^H) = \phi$.
\begin{align}
  \mathcal{C}(v_{lc}^H) \cup \mathcal{C}(v_{rc}^H) &= \mathcal{C}(v^H) \\
  \mathcal{C}(v_{lc}^H) \cap \mathcal{C}(v_{rc}^H) &= \phi
\end{align}

\begin{figure}[t]
\includegraphics[width=0.4\textwidth,height=10cm,keepaspectratio]{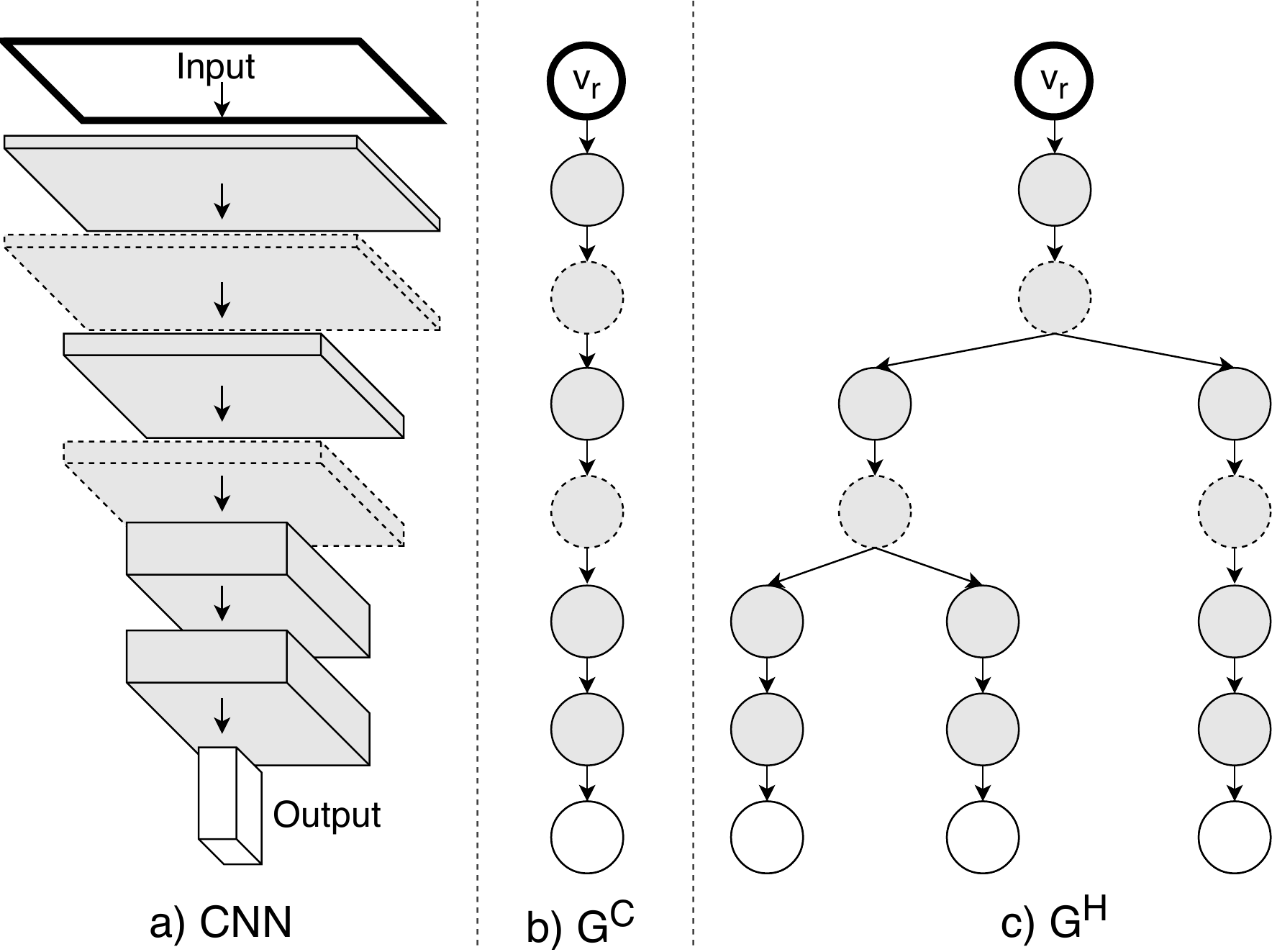}
\caption{Visualization of HSD-CNN graph formed for a specific CNN. Pre-trained CNN in a) is transformed to graph $G^C$ in b). Dotted circles signify the inclusion of pooling layers in the filters, and also a requirement of decomposing node module. Following the Algorithm \ref{al:tree}, a new HSD-CNN $G^H$ is formed. In the example shown, $G^H$ has three leaf nodes.}
\label{fig:hsd}
\end{figure}

For clustering classes into subsets, select $\textit{Iscv}^{L(v^H)}$ vectors of $v_c^C$ corresponding to classes in $\mathcal{C}(v^H)$. Clustering is performed with Ward's agglomerative clustering method \cite{doi:10.1080/01621459.1963.10500845} in a bottom-up approach. In this approach, $\textit{Iscv}(v_c^C)$ feature samples are assumed to be one cluster for each class in $\mathcal{C}(v^H)$. At each step, find those pair of clusters among them that lead to a minimum increase in total within-cluster variance, and later merge each pair as a new cluster. This increase is based on a weighted squared distance between cluster centers. The cluster distances are defined to be the squared Euclidean distance between vectors $\textit{Iscv}(v_c^C)$. Merging process is continued up to the hierarchy until we obtain two clusters. Final two clusters represents two class subsets $\mathcal{C}(v_{lc}^H), ~ \mathcal{C}(v_{rc}^H)$.

When the above clustering results in either one or $|\mathcal{C}(v^H)|-1$ cardinal number, both nodes $v_{lc}^H$ and $v_{rc}^H$ are merged to form a single node. The number of elements in the set is the cardinal number of class subsets, here.

\begin{figure*}[ht!]
\includegraphics[width=0.95\textwidth,height=10cm,keepaspectratio]{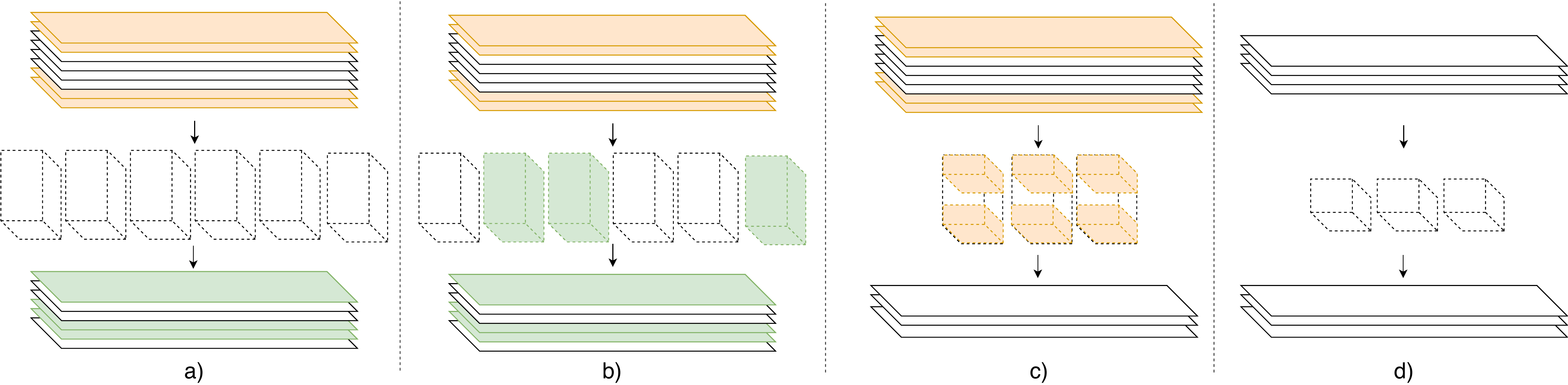}
\caption{Parameters from CNN graph $G^C$ are sequentially transferred to HSD-CNN $G^H$. Transfer process for each edge is visualized here.}
\label{fig:trans}
\end{figure*}

Though $v_c^C$, a 3-dimensional map, has $K$ channels, we select only $K/2$ channels for its decomposed nodes $v_{lc}^H, ~ v_{rc}^H$, separately. Only those $K/2$ channels are selected that have high impact scores in $\textit{Iscv}(v_c^C)$ for the classes $\mathcal{C}(v_{lc}^H)$ to form node $v_{lc}^H$. The same is repeated for $v_{rc}^H$, though the selected channels may differ.

Yet, tree structured CNN $G^H=(V^H,E^H)$ formed from Algorithm \ref{al:tree} has un-weighted edges.


\subsubsection{\textbf{Parameter decomposition - Transferring model}} \label{sss:trans}

However, designing such networks $G^H$ increases the number of parameters and nodes. Such a network cannot be optimally trained with limited samples. We require a transfer learning based approach where the initialization of the newly formed network tree $G^H$ is improved more suitably.

\begin{algorithm}[h]
\SetAlgoLined
\KwResult{Edges $E^H \in G^H$}
\tcc{Omit root node}
 \For{ each node $v^H in G^H$}{ 
 $v^C \gets$ node at $L(v^H)^{th}$ layer of $G^C$\;
 $temp \gets e(P(v^C), v^C) \in G^C$ \tcp*{K 3-d filters}
 $\mathcal{O} \gets$ set of $K/2$ channels selected for $v^H$\;
 $temp \gets temp[\mathcal{O},:,:,:]$\;
 $\mathcal{I} \gets$ set of $K/2$ channels selected for $P(v^H)$\;
 $temp \gets temp[:,:,:,\mathcal{I}]$\;
 $e(P(v^H), v^H) \in G^H \gets temp$\;
 }
\caption{Transfer edge filter parameters from $G^C$ to $G^H$}
\label{al:transfer}
\end{algorithm}

As in Algorithm. \ref{al:transfer}, for each node $v^H \in G^H$, filters $e(P(v^C), v^C) \in G^C$ are obtained, depicted in Fig. \ref{fig:trans}.a), where $L(v^C)= L(v^H)$. All the green channels of node $v^H$ and its corresponding filter channels in edge $e(P(v^C), v^C)$ are omitted as seen in Figure \ref{fig:trans}.b. Later, the $K/2$ \textit{orange} input channel maps of $P(v^H)$ and its corresponding filter parameters in edge $e(P(v^H), v^H)$ are excluded as in Figure \ref{fig:trans}.c. The truncated filters form an edge value $e(P(v^H), v^H) \in G^H$, as in Figure \ref{fig:trans}.d. Parameter transferring is repeated for all edges in $G^H$.

\subsubsection{\textbf{Training}} \label{sss:training}

Though edges $E^H \in G^H$ are transferred from $G^C$, further fine-tuning is required as the nodes position, and input to leaf nodes differ. So, the HSD-CNN network $G^H$ is further fine-tuned with the training dataset $\mathbb{D}$  

\section{Evaluation} \label{s:eval}

We implemented the decomposition of CNN and performed experiments using Pytorch library \cite{pytorch}. Training is conducted on NVIDIA GeForce 1080Ti based workstation. GPU speeds are measured on GeForce 1080Ti GPU and Intel(R) Xeon(R) E5-1660 v4 CPU. In training original network $G^C$ and decomposed network $G^H$, we use stochastic gradient descent optimizer ( SGD )\cite{DBLP:journals/corr/Ruder16}. Also, the learning rate is reduced by ten times for every 50 epochs during training with an initial learning rate of 0.01.


\subsection{Datasets}  \label{ss:dsets}
We evaluate our proposed approach on different class sizes of datasets, namely \emph{CIFAR10, CIFAR100 and CALTECH101}.

\textbf{CIFAR \cite{citeulike:7491128}}: The CIFAR dataset consists of natural images with a resolution of $32 \times 32\times3$. CIFAR10 is drawn from $10$ classes, whereas CIFAR100 consists of $100$ classes. The train and test sets in both CIFAR10 and CIFAR100 contain $50,000$ and $10,000$ images respectively.

\textbf{CALTECH101 \cite{Fei-Fei:2007:LGV:1235884.1235969}}: This is another dataset used for validating HSD-CNN for a large number of classes with higher resolution of $224 \times 224 \times 3$. It has $9155$ images containing with $101$ classes and one background category. The dataset is split into $6401$ training images, and $2744$ testing images.

In the pre-processing step, color distortion is applied to each image. Due to inconsistency in image sizes, we re-size all samples to common size using bi-linear interpolation as the images available are not consistent in their size. Further whitening effect on samples normalizes the intensity values.

\begin{table*}[ht!]
    \begin{minipage}{.48\textwidth}{
    \caption{Standard baseline statistics for VGG16 on CIFAR10, CIFAR100 and CALTECH101 datasets.}
    \label{tab:stand}
    \centering
    \resizebox{0.99\textwidth}{!}{%
    \begin{tabular}{r|r|r|r|l}
    \hline
    \textbf{DataSet} & \textbf{Accuracy} & \textbf{Parameters} & \textbf{Computations} & \textbf{Time (mSec)} \\ \hline
    CIFAR10      & 93.41 & 14.7M & 313M  & 0.246 \\
    CIFAR100     & 72.07 & 14.7M & 313M  & 0.242 \\
    CALTECH101   & 76.39 & 14.7M & 15.3G & 14.15
    \end{tabular}
    }
    }
    \end{minipage}
    \hfill
    \begin{minipage}{.48\textwidth}{
    \caption{Performance of HSD-CNN algorithm on VGG16 for CIFAR and CALTECH datasets.}
    \label{tab:res}
    \centering
    \resizebox{0.99\textwidth}{!}{%
    \begin{tabular}{r|r|r|r|r|l}
    \hline
    \textbf{DataSet}    & \textbf{Accuracy}      & \textbf{Compression}      & \textbf{SpeedUp}      & \textbf{Saved computations} & \textbf{Leaf Nodes} \\
              & \textbf{Drop}          & \textbf{Rate}             & \textbf{Rate}         & \textbf{ratio (\%)}  &   \\ \hline
    CIFAR10    & 0.08          & 1.34             & 1.51         & 32.27\% 	   & 3 \\
    CIFAR100   & 0.85          & 0.37             & 0.97         & -39.94\%    & 15 \\
    CALTECH101 & -2.56         & 0.56             & 1.82         & 1.31\%      & 10 
    \end{tabular}
    }
    }
    \end{minipage}
\end{table*}


\subsection{Networks} \label{ss:net}
We experimented the decomposition algorithm on VGG16 architecture \cite{DBLP:journals/corr/SimonyanZ14a}. VGG16 network majorly consists of 5 max-pooling layers and 13 convolutional ( CONV ) layers followed by three full connected ( FC ) layers. All the CONV layers use $3 \times 3$ filters, inclusive of batch normalization followed by a ReLU non-linear unit. Though it is possible to calculate impact scores for FC layer, it is restricted to only CONV layers. Because we either use 1 FC layer or $1 \times 1$ filtered CONV layer in combination with an adaptive global average pooling layer after CNN feature maps.

\subsection{Metrics}  \label{ss:metrics}
\textbf{Accuracy}: \emph{Accuracy} compares the top predicted class with the ground truth class, and labels them as correct if both labels are same.


$Accuracy= \#correct~labels/\#total~labels$


\textbf{Accuracy drop} is the difference between originally trained model accuracy and the model performance accuracy obtained after hierarchical decomposition algorithm.

$Accuracy~Drop = Accuracy_{Original} - Accuracy_{Decomposed}$


Assuming the number of parameters, operations and running time for a sample in original network model $M$ as $a, n, s$, respectively. Similarly, assume $a^*, n^*. s^*$ representation for decomposed model $M^*$ too. Following metrics are given by
\begin{align*}
    \textbf{Compression~Rate~\cite{DBLP:conf/iccv/ChengYFKCC15}}~~\alpha(M,M^*) =& \frac{a}{a^*}.\\
    \textbf{Saved~computations}~~\gamma(M,M^*) =& \frac{n - n^*}{n}.\\
    \textbf{Speed~Up~rate~\cite{DBLP:conf/iccv/ChengYFKCC15}}~~\delta(M,M^*) =& \frac{s}{s^*}.
\end{align*}



\subsection{Implementation and Experimental Results}  \label{ss:exp}

We chose one conventional network - VGG16 to test our algorithmic approach and perform varied experiments to demonstrate the effectiveness of our proposed approach.


    

\begin{figure*}[h]
\includegraphics[width=0.95\textwidth, height=6cm]{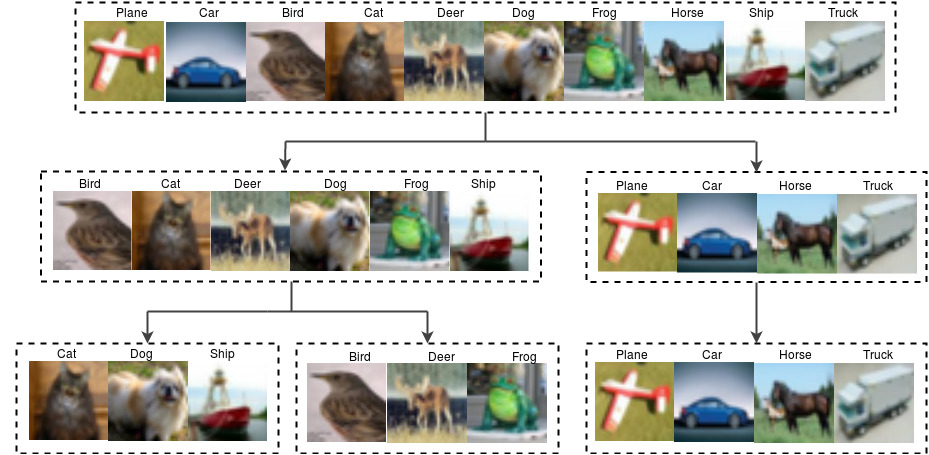}
\caption{Hierarchical representation of classes for CIFAR10 from HSD-CNN graph.}
\label{fig:hie_cifar10}
\end{figure*}


\begin{table*}[ht!]
\caption{Comparison of HSD-CNN with its leaf node sub-networks on VGG16 for CIFAR and CALTECH datasets. In the format x(y) of \textit{Accuracy}, x represents accuracy of sub-network calculated for y classes.}
\label{tab:leaf}
\centering
\resizebox{0.99\textwidth}{!}{%
\begin{tabular}{|r|r|r|r|r|l|}
\hline
\textbf{Dataset}    & \textbf{Parameters} & \textbf{Compression Rate} & \textbf{computations} & \textbf{Saved computations Ratio (\%)} & \textbf{Accuracy (\%)}                                                                                                                         \\ \hline
CIFAR10    & 10.98M              & 1.34              & 212M           & 32.27                    & 93.33                                                                                                                                     \\
           & 3.7M                & 3.98              & 89M            & 71.57                   & 93.53(3), 97.3(3), 97.73(4)                                                                                                               \\ \hline
CIFAR100   & 40M                 & 0.37                    & 438M           & -39.94                  & 71.22                                                                                                                                     \\
           & 3.7M                & 3.97               & 89M            & 71.57                   & 85.85(13), 84.9(10), 71.17(6), 85(11), 88(6), 78(6), 91.25(4), 94.75(4), 87.38(8), 72.81(16), 95.6(6), 92.6(3), 99.5(2), 98.5(2), 89.3(3) \\ \hline
CALTECH101 & 26.3M               & 0.56                      & 15.1G          & 1.31                            & 78.95                                                                                                                                     \\
          & 3.7M                & 3.97               & 4.36G          & 71.13                   & 80.1(35), 89.23(9), 91.89(17), 94.44(3), 80.57(12), 100(2), 90.63(3), 96.67(4), 97.18(6), 92.74(11)                                      \\ \hline
\end{tabular}
}
\end{table*}


\subsubsection{\textbf{Decomposing VGG16 using CIFAR and CALTECH datasets}}  \label{ssss:vgg_exp}
Let us first consider CIFAR10 dataset. We first train VGG16 network with specifications detailed in Section \ref{ss:net} for the dataset. Impact class score vectors are calculated for each layer with respect to each of the 10 classes in CIFAR10, following the Section \ref{sss:icsv}. As there are 13 CONV layers in VGG network chosen, we cluster only at $3^{rd}, ~5^{th}, ~8^{th}, ~and ~11{th}$ CONV layers into two subsets for each parent node received from its predecessor $2^{nd}, ~4^{th}, ~7^{th}, ~and~ 10^{th}$ CONV layer, respectively. We chose these layers as these are immediately followed by max-pool layers. Based on the Algorithm. \ref{al:tree} in Section \ref{sss:form}, a new HSD-CNN graph from \textit{Iscv}s of VGG16 on CIFAR10 is formed. At the final layer, we obtain 3 leaf nodes as the number of classes in CIFAR10 are only 10. Later, transfer the parameters obtained from a trained model of CIFAR10 dataset to the newly formed HSD-CNN, following the detailed strategy in Section \ref{sss:trans}. Accuracy drop after fine-tuning is $0.08\%$, almost negligible degradation in performance.

Similar experiments are repeated for other two datasets, with detailed results shown in Tables. \ref{tab:stand} and \ref{tab:res}. The proposed algorithm results in 15 and 10 leaf nodes at their final layers for CIFAR100 and CALTECH101 datasets, respectively. Yet CIFAR10 has 3 leaf nodes. Because the number of classes present in those datasets is more than 100. So, it might have been difficult to discriminate the classes at the decomposition nodes. Being with less number of classes for CIFAR10, HSD-CNN algorithm leads to less number of leaf nodes at its end layer for CIFAR10. Simultaneously, we also limit the least number of classes possibly discriminated by any node to a minimum two.

We also observe that there is only $0.85\%$ accuracy drop for CIFAR100 from its original $72.07\%$ accuracy. However, there is an improvement in accuracy for CALTECH101 from $76.39\%$ to $78.95\%$. Though training samples for CALTECH101 (\~6K) are less in comparison to CIFAR100 (~50K), CALTECH101 has 5 fewer leaf nodes. It may imply that forming a wider HSD-CNN network may not always improve the performance. Although, there are other reasons like size of input and activation maps used are different ($32 \times 32$ - CIFAR100, $224 \times 224$ - CALTECH101), information in \textit(Iscv) vectors might also affect their performance.

Experimental results in Tables. \ref{tab:stand} and \ref{tab:res} indicate that performing HSD-CNN algorithm either increase the performance or have negligible degradation in accuracy. We can further use the formed HSD-CNN to limit over-representation of networks and explore other applications, as detailed in Section \ref{ss:apl}.
 
Usually, higher the compression rate, better the compression algorithm. As we increase decomposed nodes in each layer as in Section \ref{sss:form} of HSD-CNN, the number of parameters increases gradually. The increased parameters and features might be redundant. To address this, we simultaneously prune half the channels for each node in \textit{decompose node} of Section \ref{sss:form}. This leads to one-fourth decrease of parameters for each node. As we are also pruning the parameters, our algorithm restricts the increase of parameters. As HSD-CNN for CIFAR10 is relatively thinner than CIFAR100 and CALTECH101, there is $1.34$ times of compression. As there are more number of leaf nodes in large class domain problem, the compression rate in CIFAR100 is just above half of that of CALTECH101. There is a chance of an increase in the compression rate for sub-network applications, explained in Section \ref{ss:apl}.

We also observe a computations savings of $32.27\%$ and $1.31\%$ in CIFAR10 and CALTECH101, respectively. However, we note no improvement for CIFAR100 ( $-39.94\%$). Similar performance is also observed for speedup rate metric. In summary, we prove that our HSD-CNN results in comparable performance for CIFAR10 and CALTECH101. A better improvement can also be found for CIFAR100 if there is a limit in the number of leaf nodes formed.


\begin{table}[ht!]
\caption{Comparison of our proposed HSD-CNN with other hierarchical CNN methods} \label{tab:compare}
\centering
\begin{tabular}{r|r|l}
\hline
\textbf{Dataset}            & \textbf{CIFAR10} & \textbf{CIFAR100} \\ \hline
HD-CNN \cite{yanhd}             & -                & 67.31             \\
Tree-CNN \cite{DBLP:journals/corr/abs-1802-05800}           & 86.24            & 60.46             \\
B-CNN \cite{Zhu2017BCNNBC}              & 88.22            & 64.42             \\
HSD-CNN( ours )     & 93.33            & 71.22         
\end{tabular}
\end{table}


\subsubsection{\textbf{Comparison with state of the art}} \label{sss:compare}

As seen in Table. \ref{tab:compare}, our proposed HSD-CNN performs better than other hierarchical methods \cite{yanhd,  DBLP:journals/corr/abs-1802-05800, Zhu2017BCNNBC} in both CIFAR10 and CIFAR100 datasets.

Algorithms in \cite{yanhd,  DBLP:journals/corr/abs-1802-05800, Zhu2017BCNNBC} are formed in 2-level hierarchy. And there is manual interference in forming their bottom finer level architecture. However, there is no manual interfering in HSD-CNN. The network is self-formed. Hierarchy with more than two levels is also established in between the object categories. Results from Table. \ref{tab:compare} also show better performance in our proposal for CIFAR datasets. As the network design in \cite{yanhd,  DBLP:journals/corr/abs-1802-05800, Zhu2017BCNNBC} is not fixed, it is computationally expensive and time-consuming in designing and training new networks. However, our proposed HSD-CNN algorithm automatically designs from any standard network and simultaneously loads suitable pre-initialized parameters.

It is also found that the proposed HSD-CNN performs better than the method in \cite{DBLP:conf/cvpr/GuoP17} in accuracy in almost all parameter cases.

\subsection{Application} \label{ss:apl}

First, it is easy to notice that all the classes in the dataset can inherently be represented hierarchically from the HSD-CNN structure layout. In a dataset with no category annotation, classes with similarities can be grouped in one category level, while unfamiliar classes lie in different categories. At the same time, it is easy to visualize all the classes and their parent categories in a hierarchical representation. However, HSD-CNN focuses more on how much the calculated features affect the classes, not on the similarity. Hierarchical visual of CIFAR10 from HSD-CNN is in Figure \ref{fig:hie_cifar10}. Plane, Car, and Truck are in one category. As they are formed at earlier stages, early layer features chose those classes that have a better impact in discriminating these classes. By \textit{Iscv vectors}, horse is also included in the same category.

Second, our HSD-CNN forms more than one paths. The structure layout facilitates in computing the path in parallel in different cores of CPUs or GPUs. In this way, resources can be efficiently utilized without extra allocation and overhead.

Third, as discussed earlier in Section. \ref{s:intro}, conventional CNN produces over-represented features in discriminating classes. Our HSD-CNN proposal modifies the CNN structure and can be used for limited subsets of classes which has appropriate representation of features to discriminate only those set of classes in the application. Detailed explanation in Section \ref{sss:subnet}.

\begin{figure*}[ht!]
\minipage{0.49\textwidth}
\includegraphics[width=0.99\textwidth,keepaspectratio]{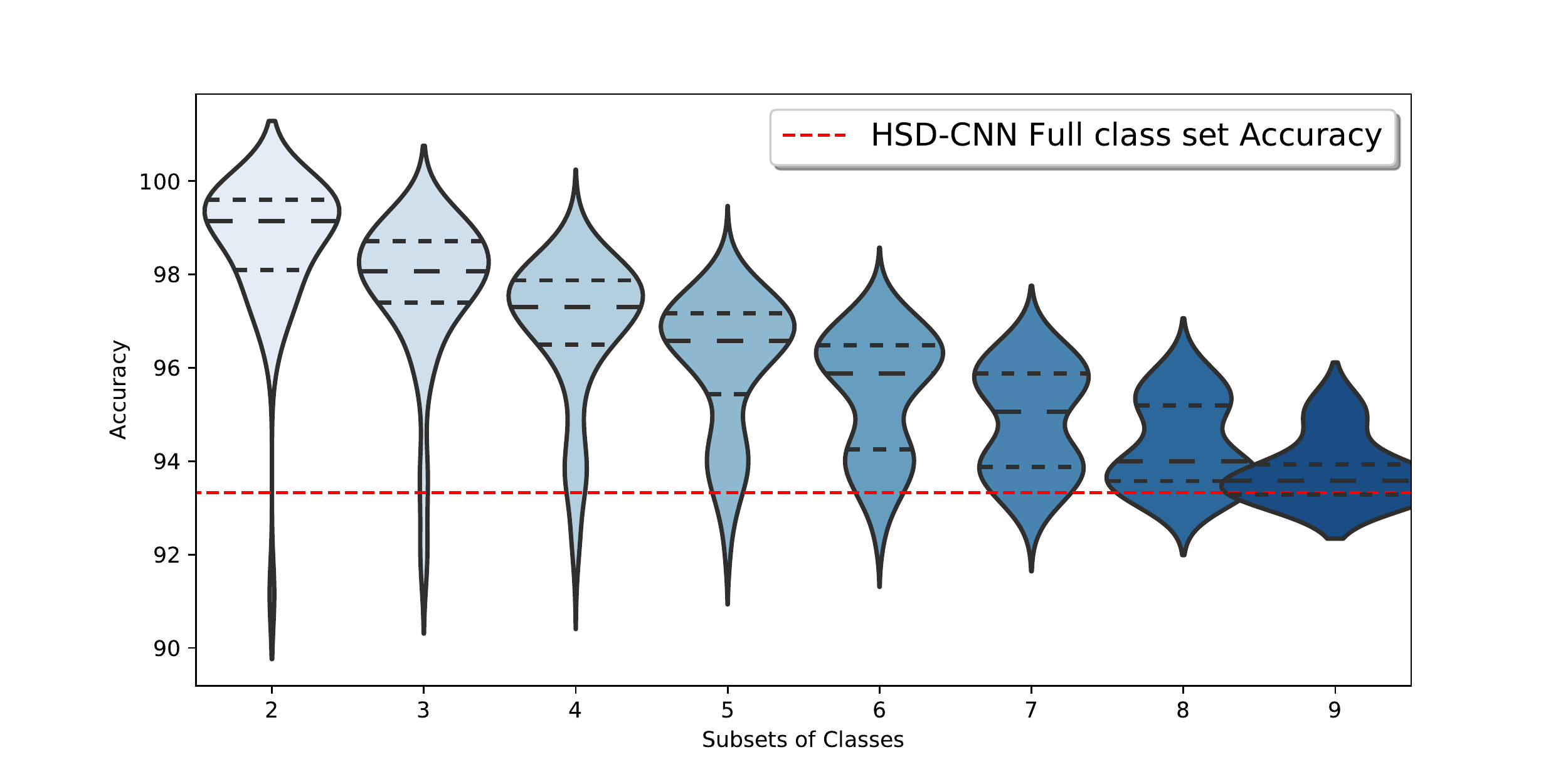}
\caption{Performance of Sub-networks formed from HSD-CNN for CIFAR10.}
\label{fig:subcifar10}
\endminipage\hfill
\minipage{0.49\textwidth}
\includegraphics[width=0.99\textwidth,keepaspectratio]{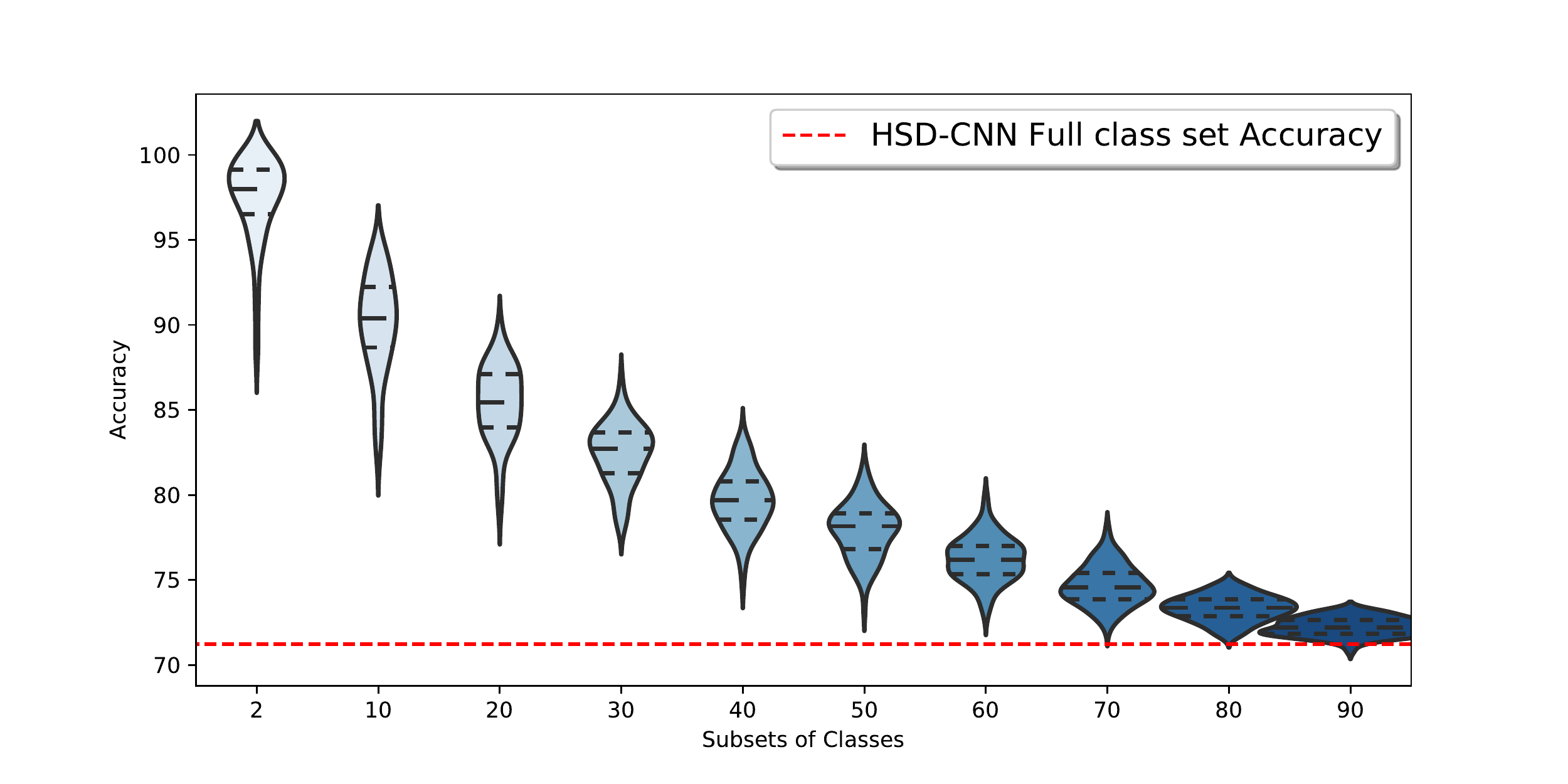}
\caption{Performance of Sub-networks formed from HSD-CNN for CIFAR100.}
\label{fig:subcifar100}
\endminipage
\end{figure*}


\subsubsection{\textbf{Sub-network}} \label{sss:subnet}

Generally, CNNs are designed and trained for hundreds or thousands of classes. These designs might prove better performance in applications of large class domains. However, all applications do not work with all the classes, and only require a subset of classes. And, it is tedious to design and train again for these applications. Our algorithm helps in retaining the performance without designing and retraining new CNN every time a different subset of a class domain is used. Experimental results from Table \ref{tab:leaf} indicate that sub-networks corresponding to leaf-nodes results in better accuracy performance. Also, these sub-networks has nearly $3.98 \times$ compression rate compared to their original network $G^C$ and saves computations by ~$71 \%$. These results are performed without retraining the sub-networks, and we are able to achieve better performance in almost all cases.

For example, HSD-CNN in CIFAR100 dataset is trained for 100 classes. Choose any 20 classes subset from its 100 classes of CIFAR100. Let us mark all the corresponding paths for these 20 classes in HSD-CNN. And form a sub-graph containing all these paths from the HSD-CNN $G^H$ with all the corresponding edges included. Now, the obtained sub-graph can be utilized directly for deployment without any further training. In this way, our HSD-CNN algorithm supports in suitable representation for limited classes, even if the network is originally trained for large class domains. Even there is no requirement of further training in subnetworks. The sub-network formed will have less number of parameters and computations, leading to an increased compression rate, speedup rate, and saved computations ratio.



\begin{figure}[ht!]
\includegraphics[width=0.45\textwidth,keepaspectratio]{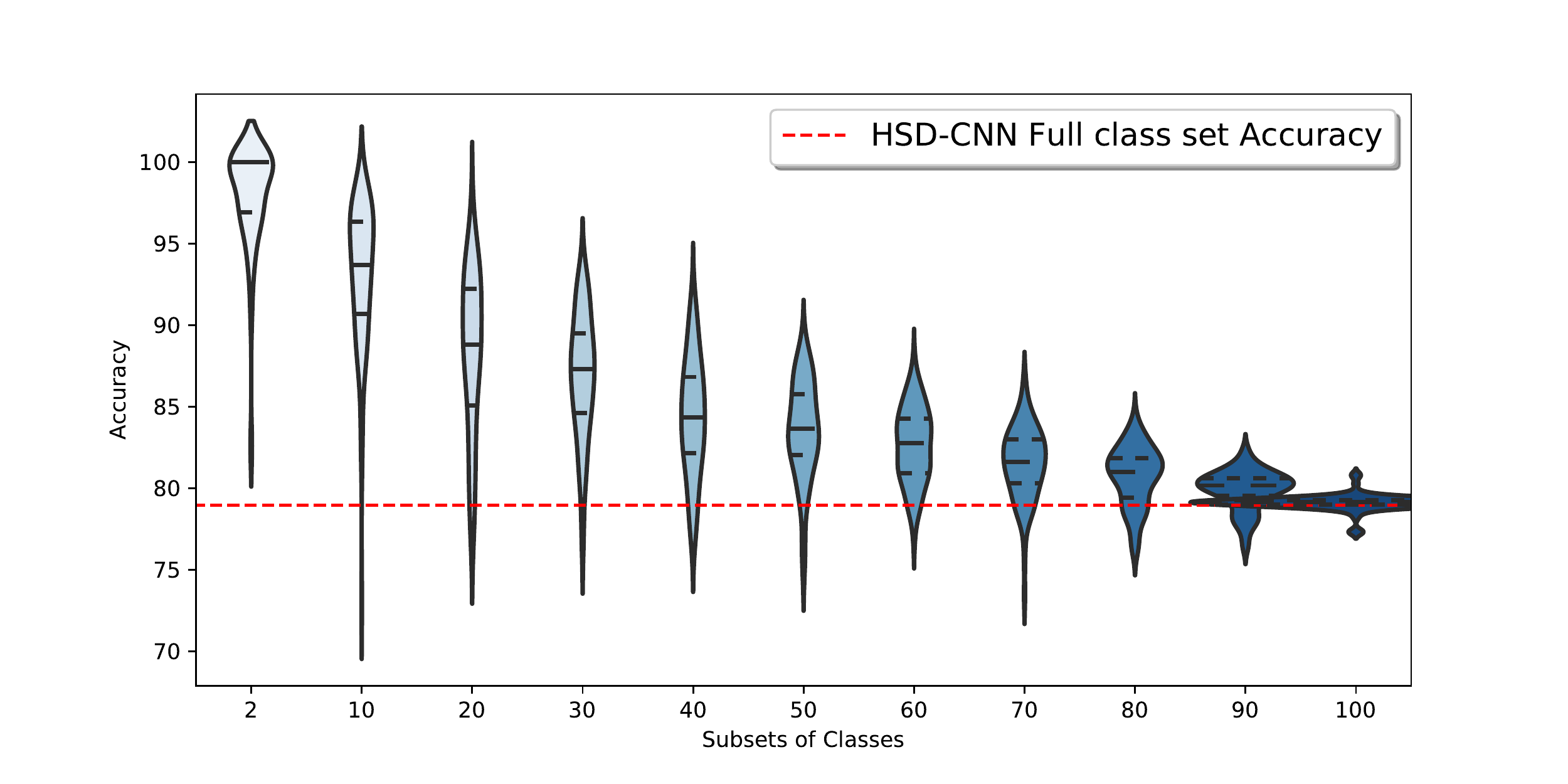}
\caption{Performance of Sub-networks formed from HSD-CNN for CALTECH101.}
\label{fig:subcaltech101}
\end{figure}

We also perform experiments to the usage of sub-networks application over HSD-CNN, as seen in Figure \ref{fig:subcifar10}, \ref{fig:subcifar100}, and \ref{fig:subcaltech101}.

\textbf{CIFAR10}: As there are $2^{10}-1$ subsets for 10 classes, we group all the combinations of classes with the same cardinal number subsets as one category ( \textit{Subsets of classes} in X-axis). We omit subset categories containing cardinal number 1 because calculating accuracy for a single class will not be valid. We perform direct inference for all these sub-networks without retraining HSD-CNN and visualize the accuracy performance in Fig. \ref{fig:subcifar10}. We also omit category with 10 cardinal number in X-axis and mark HSD-CNN accuracy for the same full class domain set as the horizontal dotted line for comparison. We observe that sub-network accuracy for the most combination of classes is higher than the full class set HSD-CNN accuracy ( dotted horizontal line ). As combinations resulting in similar accuracy, the shape at vertical line spreads more ( more points at similar accuracy ). Similarly, the thin line in the figure corresponds to those combinations of classes, which has extreme accuracy in its group. When the spread of the vertical shape is thin, it indicates there are very few combinations of classes that results in the corresponding accuracy.

We perform similar experiments for CIFAR100 and CALTECH101 datasets. However, there are nearly $2^{100}$ sub-graphs possible for these subsets. Similar experimentation requires huge memory and large time. For convenience, we chose 100 combinations of two classes for cardinal number 2. Repeat the similar 100 combinations for other cardinal numbers also. As there are nearly 100 categories on X-axis, it is difficult to visualize all on a single graph. So, we chose the cardinal numbers with multiples of 10 and category with cardinal number 2 for easier understanding of the experimentation. Even the results for other cardinal number sub-graphs follow the similar pattern observed in Figure \ref{fig:subcifar100} and \ref{fig:subcaltech101}. It is observed that sub-network performance is better their full set HSD-CNN accuracy for most cases in both datasets.


\section{Conclusion} \label{s:conc}
We propose a novel strategy to self-decompose conventional CNN in a hierarchical tree structure. We adapt class filter sensitivity analysis in calculating impact score class vectors. \textit{Iscv} vectors help in decomposition of nodes while forming HSD-CNN architecture. We also strategically initialize the newly formed HSD-CNN with parameters transferred from its original trained model. Experimental results also show that our algorithm performs better than other hierarchical methods. We also discussed sub-network applications of HSD-CNN, which limits the over-representation used in conventional CNNs. However, an increase in direct compression rate and saved computations ratio might speed up the inference time. Suitable modifications can be extended to our HSD-CNN proposal to achieve improvement in speedup rate and compression rate. As the structure of HSD-CNN looks like a tree, deciding the path of the input sample computed would definitely enhance the speedup rate and energy savings, which we intend to explore in future.

\bibliographystyle{unsrt}
\bibliographystyle{ACM-Reference-Format}

\end{document}